\documentclass[article]{IEEEtran}
\usepackage{cite}
\usepackage{amsmath,amssymb,amsfonts}
\usepackage{algorithmic}
\usepackage{graphicx}
\usepackage{algorithm,algorithmic}
\usepackage{hyperref}
\hypersetup{
    colorlinks=true,
    allcolors=black
}
\usepackage{textcomp}
\def\BibTeX{{\rm B\kern-.05em{\sc i\kern-.025em b}\kern-.08em
    T\kern-.1667em\lower.7ex\hbox{E}\kern-.125emX}}
\usepackage{generic}

\usepackage{amsmath}

\allowdisplaybreaks
\usepackage{xcolor}
\usepackage{footnote}
\usepackage{algorithm}
\usepackage{algorithmic}
\usepackage{multirow} 
\usepackage{booktabs}
\usepackage{graphicx}
\usepackage{amssymb}
\usepackage{amsbsy}

\usepackage{amsthm}

\usepackage{array}
\usepackage{longtable}
\usepackage{epstopdf}
\usepackage{pbox}
\usepackage{breqn}
\usepackage{mathrsfs}
\usepackage{multicol}
\usepackage{supertabular}
\usepackage{enumerate}
\usepackage[colorinlistoftodos]{todonotes}
\usepackage{url}
\usepackage[compatibility=false]{caption}
\usepackage{tabu}
\usepackage{subcaption}
\usepackage{graphics} % for pdf, bitmapped graphics files
\usepackage{epsfig} % for postscript graphics files

\usepackage{mathtools}

\usepackage{tikz}
\usepackage{tikz-network}

\usepackage{cases}
\usepackage{bm}
\usepackage{cite}

\usepackage{aligned-overset}

\newtheorem{thm}{Theorem}
\newtheorem{lem}{Lemma}

\newtheorem{assum}{Assumption}

\theoremstyle{definition}

\newcommand{\thistheoremname}{}
\newtheorem*{genericthm*}{\thistheoremname}
\newenvironment{namedthm*}[1]
  {\renewcommand{\thistheoremname}{#1}%
  \begin{genericthm*}}
  {\end{genericthm*}}

\usepackage{cite}
\RequirePackage{etex}
\usepackage{amsmath}
\begin{document}
\title{
Semi-Gradient SARSA Routing with Theoretical Guarantee on Traffic Stability and Weight Convergence
}

\author{Yidan Wu, Yu Yu, Jianan Zhang, \IEEEmembership{Member, IEEE}, and Li Jin, \IEEEmembership{Senior Member, IEEE}
\thanks{This work was in part supported by NSFC Project 62473250, SJTU-UM Joint Institute, and J. Wu \& J. Sun Endowment Fund.}
\thanks{Y. Wu is with the UM Joint Institute, Shanghai Jiao Tong University, Shanghai 200240, China (email: wyd510@sjtu.edu.cn).}
\thanks{Y. Yu is with the School of Computer Science, Wuhan University, Hubei 430072, China (email: Yu.Yu1024@whu.edu.cn).}
\thanks{J. Zhang is with the School of Electronics, Peking University, Beijing 100871, China (email: zhangjianan@pku.edu.cn).}
\thanks{L. Jin is with the UM Joint Institute and the Department of Automation, Shanghai Jiao Tong University, Shanghai 200240, China (email: li.jin@sjtu.edu.cn).}
}

\maketitle

\begin{abstract}
% no more than 300 words
We consider the traffic control problem of dynamic routing over parallel servers, which arises in a variety of engineering systems such as transportation and data transmission. We propose a semi-gradient, on-policy algorithm that learns an approximate optimal routing policy. The algorithm uses generic basis functions with flexible weights to approximate the value function across the unbounded state space. Consequently, the training process lacks Lipschitz continuity of the gradient, boundedness of the temporal-difference error, and a prior guarantee on ergodicity, which are the standard prerequisites in existing literature on reinforcement learning theory. To address this, we combine a Lyapunov approach and an ordinary differential equation-based method to jointly characterize the behavior of traffic state and approximation weights. Our theoretical analysis proves that the training scheme guarantees traffic state stability and ensures almost surely convergence of the weights to the approximate optimum. We also demonstrate via simulations that our algorithm attains significantly faster convergence than neural network-based methods with an insignificant approximation error.

\end{abstract}


\section{Introduction}\label{intro}

\subsection{Motivation}
%Background

Dynamic routing is a classical control problem in transportation, manufacturing, and networking \cite{mitra1991comparative,down1997piecewise,alanyali1997analysis}. Classical dynamic routing schemes rely on Lyapunov methods to study traffic stability and provide a solid foundation for our work \cite{kumar1995stability,dai1995stability,Jin2018Stability}.
However, these methods are less powerful to search for routing policies that optimize average system time. This problem was conventionally challenging, because analytical characterization of the steady-state distributions of the traffic state and thus of the long-time performance metrics (e.g., queuing delay) is very difficult \cite{dai2022queueing,xie2022stabilizing}. In particular, existing results usually assume non-idling policies, which can be quite restrictive. 

\begin{figure}[htbp]
        \centerline{\includegraphics[width=1\linewidth]{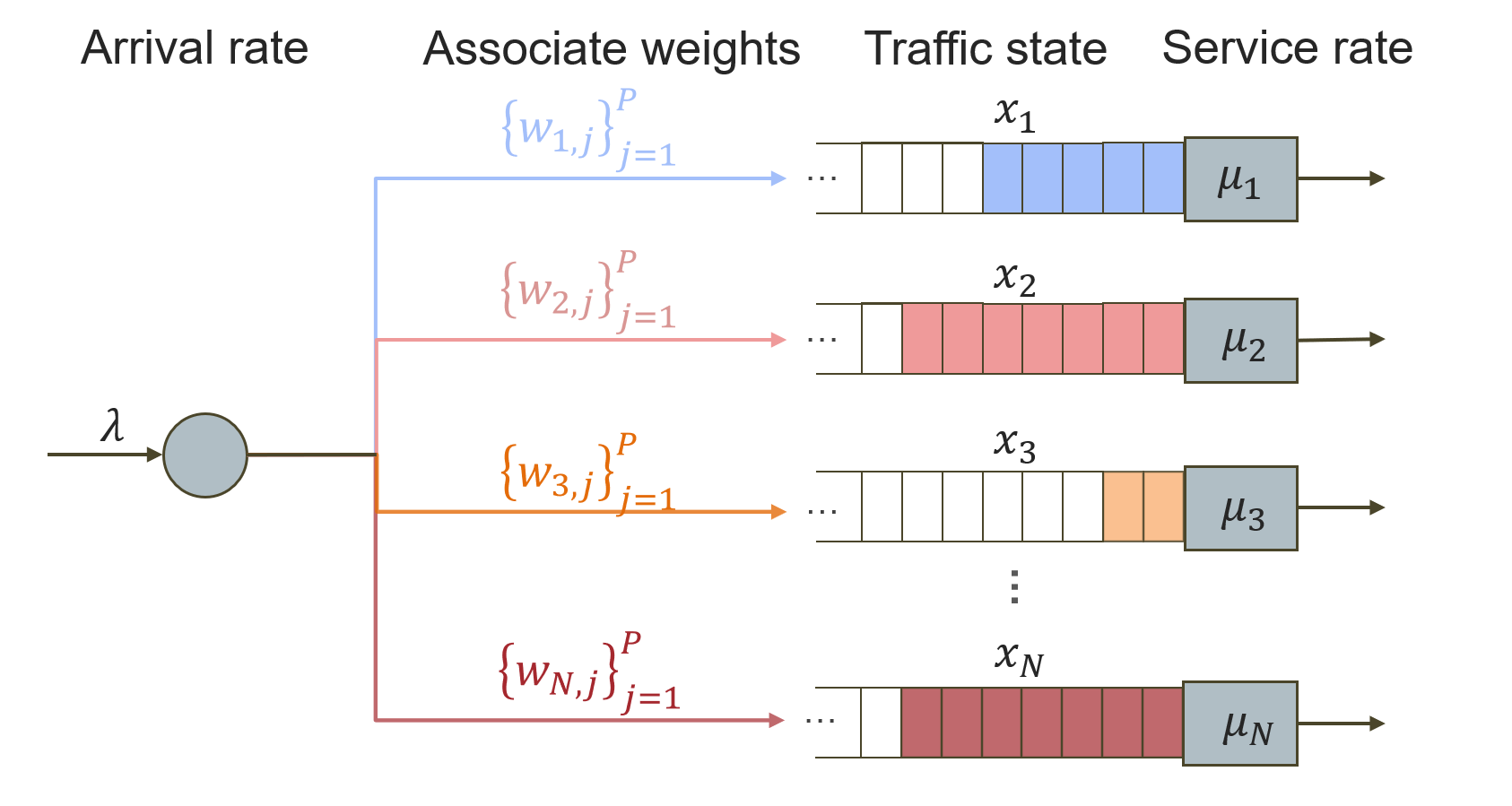}}
    \caption{A parallel service system.}
    \label{fig:parallelQN}
\end{figure}

Recently, there has been a rapidly growing interest in applying reinforcement learning (RL) methods to dynamic routing and network control problems in general. RL methods are attractive because of their  computational efficiency  and adaptivity to unknown/non-stationary environments \cite{sutton2018reinforcement}. However, there is still a non-trivial gap between the demand for theoretical guarantees on key performance metrics and the black-box nature of RL methods. In the domain of control problems, the evaluation of RL performance mostly relies on empirical approaches, thus often lacking formal guarantees concerning the reliability and robustness of the derived control policies \cite{HanZhang20}.

%Main questions
In this paper, we try to address the above challenge by studying the behavior of a parallel service system (Fig.~\ref{fig:parallelQN}) controlled by a class of semi-gradient SARSA (SGS) algorithms with generic linear function approximation; these methods are attractive because of (i) adaptivity to unknown model parameters; (ii) potential to obtain near-optimal policies and (iii) theoretical tractability.
The specific research questions are:
\begin{enumerate}
    \item How to implement RL algorithm for policy learning in dynamic routing?
    \item Under what conditions does the proposed SGS algorithm ensure the joint convergence of the weights and the state?
\end{enumerate}

\subsection{Related Work}
Tabular algorithms like Q-learning and SARSA \cite{sutton2018reinforcement} offer simplicity and intuitive understanding with strict theoretical guarantees. 
To the context of dynamic routing, \cite{liu2022rl} proposed a mixed scheme that uses learning over a bounded set of states and uses a known stabilizing policy for the other states; \cite{Camelo20} used state partition method and parallel RL to update Q-table and improve learning efficiency. However, tabular algorithms are facing the problem of `curse of dimensionality' and low computation efficiency. Though these researches extend tabular RL to a wider state space, they still focus on the theoretical guarantee of the local compact state space and lack global explainability; the mixed policy may also cause the problem of inconsistency and instability. Thus, tabular RL is not enough for dynamic routing, since the stability guarantee in unbounded state space is an important issue in dynamic routing.

While deep RL (DRL) algorithms are prevalent for their efficiency, they have poor interpretability. An important method to tackle the problem of the `curse of dimensionality' of RL algorithm is using function approximation. Motivated by the development of deep learning, people introduced neural network (NN) to RL. With the powerful and complicated NN as the approximator for value of large state space, many DRL algorithms outperform tabular RL algorithm \cite{hoang2023deep,GuptaDRL}. Among this, a class of popular algorithm Proximal Policy Optimization (PPO) \cite{schulman2017proximal} is overwhelming recently for its state-of-the-art performance in realistic systems, such as dynamic routing \cite{ChenMAPPO,dai2022queueing}. However, its off-policy nature combined with a black-box policy network not only reduces explainability but also raises safety concerns for the control policy \cite{Quentin24NUDGE,Huan22}, due to issues such as distribution mismatch, high variance in importance sampling, and sensitivity to hyperparameters. 

Among this, RL with linear function approximation \cite{Heeswijk2017TheDD,ULMER2020183}
strike a highly favorable compromise between tabular RL and DRL. It inherits certain theoretical assurances akin to those of tabular RL, and endows RL algorithms with the ability to scaling to larger and more complex environments. 
As for the convergence of RL algorithm with linear function approximation, there are two broad methods. i) Ordinary differential equation (ODE) method: It is proved in \cite{de2000existence} that the greedy RL can be represented as an ODE with a fixed point; the solution of ODE is exactly the optimal solution of Bellman equation (BE). Many researchers derived the conclusion of asymptotic convergence by analyzing the trajectory of associated ODE \cite{Karmakar21,yaji2019analysis}. In particular, \cite{melo2008analysis} proved the almost surely convergence of Q-learning and SARSA with compact state space by using ODE method. ii) Contraction scheme method: With the help of Robbins-Moron theorem \cite{robbins1951stochastic}, many researchers use contraction scheme to analyze the iteration process \cite{Qu20,CHEN2022110623,chen2024lyapunov}. \cite{Tsitsiklis97} analyzed the a.s. convergence of TD prediction with finite state space by contraction scheme and provided some insights on error analysis.

The asymptotic convergence of RL with compact state space or bounded basis function and finite sample convergence of RL have been thoroughly researched. However, existing theories mostly consider RL and MDP with the following assumptions: i) compact state space or bounded feature function; ii) bounded reward and iii) regularity and ergodicity of MDP. Many researchers have been working on extending the theory and loosening the above assumptions. In particular,  \cite{shah2020stable,grosof2024convergence} proposed convergence criteria for RL policies with unbounded state space but bounded reward. Our former work \cite{cdc24} loosed the above assumptions with a set of quadratic basis function and but got only a weak convergence conclusion. 
Note that the assumptions of ergodicity (i.e., existence of an unique steady-state distribution) and bounded TD error may be violated in dynamic routing: The jobs are arriving continuously and the iterations of the policy and system state are closely coupled. That is, the TD error may be unbounded with unbounded state space and reward; a general RL may learn a poor policy and lead to the irregularity of the MDP. Conversely, a stable system provides reliable data for policy optimization, while a well-converged policy can in turn enhance system stability. Thus, there is still a lack of solid theory on the convergence of value function approximation over unbounded state spaces, basis functions, rewards and more specific learning algorithms for dynamic routing; such a theory and learning algorithm are essential for developing interpretable and reliable routing schemes.

\subsection{Methodology and Contribution}
%Modeling
In response to the above research gaps, we develop a on-policy reinforcement learning algorithm based on insights for traffic control.
The routing objective is to minimize the expected total system time in the parallel service system illustrated in Fig.~\ref{fig:parallelQN}, which includes waiting times and service times, thus maximizing the expected utility.
We use linear combinations of a set of basis functions to approximate the action value function, since analytical solution is not easy.
The weights parameterize the approximate function and thus determine the routing policy. 
In particular, appropriate choice of basis functions naturally makes possible idling policies, which is not always the case with many existing methods. The weights are updated by a semi-gradient on-policy algorithm. 

%Results
The main result (Theorem~\ref{thm: joint convergence}) states that the proposed algorithm ensures joint convergence of the traffic state (in the sense of ergodicity and bounded norm) and the weight vector (in the sense of almost surely) if and only if the system is stabilizable. This result requires only rather mild assumptions. Importantly, we study the coupling between the long-time behavior of the traffic state and that of the weight vector, which extends the existing theory on finite-state problems to our setting with unbounded state spaces and unbounded reward. The convergence of traffic state is established by showing that a Lyapunov function associated with the approximate value function verifies the drift criterion \cite{meyn_tweedie_1993}. The convergence of weights is established based on stochastic approximation theory \cite{Benveniste90}. 
 
We validate the proposed algorithm by demonstrating an example in TCP network congestion control. This problem involves a nonlinear cost function. 
We compare the proposed algorithm with the classical join-the-shortest-queue (JSQ) algorithm and a representative neural network-based SARSA (NNS) algorithm. Our algorithm gives an empirical average cost 41\% less than that given by JSQ. More importantly, our algorithm converges over 5 times faster than NNS (which has no theoretical guarantees on stability or convergence), with a rather insignifant optimality gap of 2\%.

%Contributions
In summary, the contributions of this paper are as follows.
\begin{enumerate}
    \item We propose a structurally intuitive algorithm to learn near-optimal routing policies over parallel servers. 

    \item We study joint convergence of traffic state and weight vector under the proposed algorithm.

    \item We present empirical evidence for the computational efficiency and near-optimality of the proposed algorithm.
\end{enumerate}

%Paper structure
The rest of this paper is organized as follows. Section~\ref{sec_model} introduces the parallel service system model, the MDP formulation, and the SGS algorithm. Section~\ref{sec_stable} presents and develops the main results on learning algorithm and joint convergence. Section~\ref{sec_exp} compares the SGS algorithm with two benchmarks. Section~\ref{sec_con} gives the concluding remarks.
\section{Modeling and Formulation}
\label{sec_model}

Consider the system of parallel servers with infinite buffer sizes in Fig.~\ref{fig:parallelQN}. In this section, we model the dynamics of the system, formulate the dynamic routing problem as a Markov decision process (MDP), and introduce our semi-gradient SARSA (SGS) algorithm, which is designed to efficiently solve the routing problem under the given system constraints.
\subsection{System modeling}\label{subsec_SPQN}
Let $\mathcal{N}=\{1,2,3,\ldots,N\}$ represent the set of $N$ parallel servers in the system. Each server $n$ has an exponentially distributed service rate $\mu_n$ and handles $x_n(t)$ jobs at time $t \in \mathbb{R}_{\geq 0}$. The state of the system is $x=[x_1,x_2,\ldots,x_N]^T$, and the state space is $\mathbb{Z}^N_{\geq 0}$. Jobs arrive at origin $S$ according to a Poisson process of rate $\lambda> 0$.

Routing actions are taken only at transition \textit{epochs} \cite[p.72]{gallager2013stochastic}. Therefore, the routing problem in the parallel queuing system can be represented as a discrete-time (DT) MDP with a countably infinite state space $\mathbb{Z}^N_{\geq 0}$ and a finite action space $\mathcal{N}$. With a slight abuse of notation, we denote the state and action of the DT MDP as $x[k]\in\mathbb{Z}^N_{\geq 0}$ and $a[k]\in\mathcal{N}$, respectively. Specifically, $x[k]=x(t_k)$, where $t_k$ is the $k$-th transition epoch of the continuous-time process. As discussed in Section~\ref{subsec_SPQN}, the routing policy is parameterized by a weight vector $w \in \mathbb{R}^N$, which influences the decision-making process across servers. 

The transition probability $\mathsf p(x'|x,a)$ of the DT MDP is directly derived from the defined system dynamics. Let $e_i\in\{0,1\}^N$ denote the unit vector, where $e_{i,i}=1$ and $e_{i,j}=0,j\not=i$, to represent the basis vectors of the state space. Then for $\forall a,n\leq N$, we have
\begin{align*}
    \mathsf p(x'|x,a)=\begin{cases}
        \frac{\lambda}{\lambda+\sum_{n=1}^N\mu_n\mathbb{I}_{\{x_n>0\}}} &x'=x+e_a,\\
        \frac{\mu_n\mathbb{I}_{\{x_n>0\}}}{\lambda+\sum_{n=1}^N\mu_n\mathbb{I}_{\{x_n>0\}}}&x'=x-e_n,x_n\geq1.
    \end{cases}
\end{align*}

We say that the traffic in the system is stochastically bounded if there exists a constant $M<\infty$ such that for any initial condition,
\begin{align}
    \label{equa: stable state}
    \lim_{t\rightarrow\infty}\frac{1}{t}\int_{s=0}^t\mathsf E[\|x(s)\|_1]ds<M,
\end{align}
where $\|\cdot\|_1$ is the standard $1$-norm for $\mathbb{R}^N$. 
We say that the traffic in the system is convergent in distribution if there exists a unique invariant probability measure $d^*(\cdot)$ on $\mathbb Z^N_{\ge0}$ such that for every initial condition $x_0$,
\begin{align}
\label{equa: TV converge}
    \lim_{n\rightarrow\infty}\|\mathsf p^n(x_0,\cdot)-d^*(\cdot)\|_{\mathrm{TV}}=0,
\end{align}
where $\|\cdot\|_{\mathrm{TV}}$ denotes the total variation (TV) norm and $\mathsf p^n(x_0,\cdot)$ denotes the 
$n$-step transition kernel \cite{meyn2012markov}.
We say that the system is \textit{stabilizable} if 
\begin{align}
    \label{equa: stabilizable}
    \lambda<\sum_{n=1}^N\mu_n.
\end{align}
Note that the above ensures the existence of at least a stabilizing Bernoulli routing policy.

\subsection{MDP formulation}

When a job arrives, it will go to one of the $N$ servers according to a routing policy 
$$
\pi: \mathcal{N}\times \mathbb{Z}^N_{\geq 0}\rightarrow [0,1].
$$
That is, $\pi(a|x)$ is the probability of routing the new job to server $a$ conditional on state $x$. 

The one-step random cost of the MDP, representing the immediate cost incurred at each transition epoch, is given by 
\begin{align*}
    c[k+1]=\sum_{n=1}^N C_n(x'_n)\cdot(t_{k+1}-t_k),
\end{align*} 
where $x'=x[k+1],C_n:\mathbb{Z}_{\geq 0}\to \mathbb{R}_{\geq 0},n=1,2,\ldots,N$ is a set of non-linear functions with engineering interpretations \cite{gurvich2009scheduling,mandelbaum2004scheduling,dai2008optimal}.
The expected value of the one-step random cost, conditioned on the current state and action, is defined as
\begin{align*}
    \bar c(x,a)=\sum_{x'}\mathsf p(x'|x,a)\sum_{n=1}^N C_n(x'_n)\frac{1}{\lambda+\sum_n \mu_n\mathbb{I}_{\{x_n>0\}}}.
\end{align*}

The state-action value for the infinite-horizon process is thus given by
\begin{align*}
    &Q_\pi(x,a)=\\
    &\quad\mathsf{E}_\pi\Big[\sum^\infty_{k=0}\gamma^{k}\sum_{n=1}^NC_n(x_n[k+1])(t_{k+1}-t_{k})\Big|x,a\Big],
 \end{align*}
where $x[0]=x,a[0]=a$ and $\gamma\in(0,1)$ is the discount factor.
Note that the ergodicity of the process needs to be proved to ensures the existence of expectation $\mathsf E_{\pi}[\cdot]$, which is, however, is typically assumed by default in most convergence analyses \cite{melo2008analysis,Tsitsiklis97}.
Let $Q^*(x,a)$ denote the solution of Bellman optimal equation
\begin{align*}
    Q^*(x,a)=\bar c(x,a)+\gamma\min_{a'}\sum_{x'}\mathsf p(x'|x,a)Q^*(x',a'),
\end{align*}
and let $\pi^*$ denote the greedy policy with respect to $Q^*$.

Closed-form solution to $Q_\pi$ is not easy. Therefore, we approximate $Q_\pi$ using
\begin{align}
    \label{equa:def Q}
    &\hat Q(x,a;w)=\sum_{n=1}^N \sum_{j=1}^{P}w_{n,j}\phi_{n,j}(x_n+\mathbb{I}_{\{n=a\}}),
\end{align}
where $w=[w_{1,0},\ldots,w_{1,P-1},\ldots,w_{N,0},\ldots,w_{N,P-1}]^T \in\mathbb{R}^{NP}$ is the weight vector of the parallel queues; $\phi_{n,j}:\mathbb{Z}_{\geq 0}\to\mathbb{R}_{\geq0}$ is a generic value function for $x_n$ and $\{\phi_{n,j};1\leq n\leq N,1\leq j\leq P\}$ are $(NP)$ linearly independent functions. 

For each $n\in\mathcal{N}$, we use $\phi_{n,H}(x_n), H\leq P$ to denote the basis function with the highest-degree of $x_n$. Let $\phi_n(x_n) = \sum_{j=1}^{P}\phi_{n,j}(x_n)$ denote the summation of associated basis value functions. Specifically, the generic functions $\phi_{n,j}(x_n),\forall n\leq N,j\leq P$ satisfy the following assumption. 
\begin{assum}
\label{assum: basis func}
    For every $n\in\mathcal{N},x_n\in\mathbb{Z}_{\geq0}$, the summation of associated basis value function satisfies $\phi_n(x_n) =\mathcal{O}(C_n(x_n))$ and $\lim_{x_n\to\infty}\phi_n(x_n)\to\infty$. $\phi_{n,j}(x_n)$ satisfies the following conditions.
    \begin{enumerate}
        \item The basis function $\phi_{n,j}(x_n)$ is non-decreasing.
        \item The first, second and third order derivatives $\phi'_{n,H}(x_n),\phi''_{n,H}(x_n),\phi'''_{n,H}(x_n)$ exist and there exists a finite positive constant $B_{de}$ such that for every $x\in\mathbb{Z}^N_{\geq0},1\leq n\leq N$,
        \begin{align*}
            &\phi''_{n,H}(x_n)\leq B_{de},\\
            &\phi'''_{n,H}(x_n)\leq B_{de}.
        \end{align*}
        \item There exist $\epsilon_w>0$ and a positive constant $B_l$ such that for $x_n\geq B_{l}$  
        \begin{align*}
            \Big(\frac{\lambda}{\sum\mu_n}-1\Big)\phi'_{n,j}(x_n)+4B_{de}\leq -\epsilon_w.
        \end{align*}
    \end{enumerate}
\end{assum}
Note that this assumption can be easily satisfied by a board class of basis functions, e.g. polynomial basis functions, logarithm basis function, etc., under the condition of \eqref{equa: stabilizable}.

With the approximation of the action-state value function $\hat Q: \mathcal{N}\times \mathbb{Z}^N_{\geq 0}\times \mathbb{R}^N\to \mathbb{R}_{\geq 0}$ defined as \eqref{equa:def Q}, the softmax policy can be explicitly defined as
\begin{align}
\label{equa: policy}
    \pi_{w}(a|x)=\frac{\exp(-\hat Q(x,a;w)/\iota)}{\sum_{b=1}^{N}\exp(-\hat Q(x,b,w)/\iota)},
\end{align}
where $\iota\in(0,\infty)$ is the ``temperature''. Note that $\pi_w(a|x)$ approaches a deterministic policy greedy with respect to $ \hat Q$ as $\iota\to0$ \cite{singh2000convergence}.

With the definition of weight $w$ and associated policy $\pi_w$, we define the drift operator $\mathcal{L}_w$ for $f(x):\mathbb{R}^N\mapsto [0,\infty)$ as
\begin{align*}
    \mathcal{L}_w f(x)=\sum_a\pi_w(a|x)\sum_{x'}\mathsf p(x'|x,a)\Big(f(x')-f(x)\Big).
\end{align*}

Let $w^*$ denote the optimal solution to
\begin{align}
\label{equa: def w^*}
    \min_w\sum^N_{\substack{x\in\mathbb{Z}^N_{\geq0}, a\leq N}}d^*(x)\pi^*(a|x)\Big(Q^*(x,a)-\hat Q(x,a;w)\Big)^2.
\end{align}

\subsection{Learning algorithm}
\label{subsec: sgs algorithm}
Suppose that the lower bound of the element of weight $w_{n,H}$ is $w_l>0$, that is, there exist $\epsilon_l>0$ that satisfies $w_{n,H}\geq w_l+\epsilon_l,\forall n=1,2,\ldots,N,H\leq P$. Let $w[k]$ denote the weight vector at the $k$-th transition epoch, which is updated by SARSA(0) algorithm
\begin{align*}
    &w[k+1]\\
    &=w[k]+\frac{1}{B_\alpha[k]}\alpha_k\Delta[k]\nabla_w\hat Q(x[k],a[k];w[k]);
\end{align*}
in the above, $B_\alpha[k]\in[1,\infty),\forall k\in\mathbb{Z}_{\geq0}$ are restraint parameters which guarantee the lower bound and continuous update of the weight, $\alpha_k$ is the stochastic step size, $B_\alpha[k]$ is the constrained parameter, $\Delta[k]$ is the temporal-difference (TD) error, and $\nabla_w\hat Q(x[k],a[k];w[k])$ is the gradient, which are specified as follows.

Define the vector valued function
\begin{align*}
    \phi(x,a)&=\Big[\phi_{1,1}(x_1+\mathbb{I}_{\{a=1\}}),\phi_{1,2}(x_1+\mathbb{I}_{\{a=1\}}),\\
    &\quad\ldots,\phi_{N,P}(x_N+\mathbb{I}_{\{a=N\}})\Big]^T\in\mathbb{R}^{NP}_{\geq0}.
\end{align*}
Then we can compactly write
\begin{align*}
    &\hat Q(x,a;w)=w^T\phi(x,a),\\
    &\triangledown_w\hat Q(x,a;w)=\phi(x,a).
\end{align*}
Then, for any $k\in\mathbb{Z}_{\geq 0}$, the TD error and the gradient are collectively given by
\begin{align}
\label{equa: def TDerror}
    &\delta_{w[k]}(x[k],w[k])
    =\Delta[k]\nabla_w\hat Q(x[k],a[k];w[k])\nonumber\\
    &=
    \Bigg(-\phi^T\Big(x[k],a[k]\Big)w[k]+c[k+1]\nonumber\\
    &\qquad+\gamma\phi^T\Big(x[k+1],a[k+1]\Big)w[k]\Bigg)\phi\Big(x[k],a[k]\Big)\nonumber\\
    &=g_{w[k]}(x[k])\cdot w[k]+r_{w[k]}(x[k]).
\end{align}

We assume that the step size sequence satisfies the following assumption
\begin{assum}The sequence $\{a_k\},k\in\mathbb{Z}_{\geq0}$ satisfies 
\label{assum:eq_sumk=0}
    \begin{align*}
    \sum_{k=0}^\infty\alpha_k=\infty,
    \quad
    \sum_{k=0}^\infty\alpha^2_k<\infty.
\end{align*}
\end{assum}
That is, the step sizes can not be too small to stop the iteration process, while also can not be too large to impede convergence. 

For convenience, we use $\phi_{n,j}(x,a)$ denote the $(nP+j-P)$-th element of $\phi(x,a)$.
Recall that the weights of the basis function with the highest order for each $x_n$ are lower bounded by $w_l$, then the restraint parameters are defined as
\begin{align*}
    B_\alpha[k]=\max_{n}\left\{\frac{\alpha_k\Delta[k]\phi_{n,H}(x[k],a[k])}{w_l-w_{n,H}[k]},1\right\}.
\end{align*}
When $w=w^*$, there is $B_\alpha[k]=1$.

The update equation thus becomes
\begin{align}
    w[k+1]=w[k]+\frac{1}{B_\alpha[k]}\alpha_k\delta_{w[k]}(x[k],w[k]).
    \label{equa: update w SARSA0}
\end{align}
We say that the SGS algorithm is convergent if there is
\begin{align*}
    \lim_{k\rightarrow\infty}w[k]=w^*, a.s.,
\end{align*}
for every initial traffic state $x[0]\in\mathbb{Z}_{\geq 0}^N$, action $a[0]\in\mathcal{N}$ and every initial weight $w[0]\in\mathbb{R}^N, w_{n,H}[0]\geq w_l,n\leq N$, where $\|\cdot\|_2$ is the standard 2-norm for $\mathbb{R}^N$.

\section{Joint convergence guarantee}\label{sec_stable}
In this section, we develop the main result of joint convergence, which states that with an appropriate fine-tune parameter $\{\alpha\}$ and $\gamma$, the proposed semi-gradient SARSA (SGS) algorithm ensures joint convergence of traffic state and weight vector if and only if the parallel service system is stabilizable.
\begin{thm}(Joint convergence) Consider a parallel service system with arrival rate $\lambda>0$ and service rates $\mu_1,\mu_2,\ldots,\mu_N>0$ which satisfy $\lambda<\sum_{n=1}^N\mu_n$. Suppose that Assumptions~\ref{assum: basis func} and \ref{assum:eq_sumk=0} are satisfied.
Then, the semi-gradient SARSA algorithm ensures that the traffic state process $\{x[k]\}$ is convergent in the sense of \eqref{equa: stable state} and \eqref{equa: TV converge} and the weight training process $\{w[k]\}$ converges to the optimal solution $w^*$ to \eqref{equa: def w^*} a.s.
\label{thm: joint convergence}
\end{thm}

To wield the ODE approach to establish the joint-convergence, we first prove that the weighted shortest queue policy and SGS algorithm naturally satisfy the required assumptions; then we are able to use ODE method to analyze the asymptotic convergence and prove Theorem~\ref{thm: joint convergence}. According to Assumption~\ref{assum:eq_sumk=0}, the step size sequence satisfies the standard condition, ensuring convergence in stochastic approximation. Then Theorem \ref{thm: joint convergence} can be analyzed with the following three steps: i) Stochastic boundedness and ergodicity: By proposing Lemma~\ref{lem: bounded and ergodic}, we verify that for every fixed $w$ during iteration, the Markov chain induced by the policy $\pi_w$ is ergodic with a unique invariant state distribution. Moreover, the moment generating function (MGF) of the states exists and is bounded, ensuring the Markov chain's well-behavedness. ii) Lipschitz continuity: By proposing Lemma \ref{lmm:Lip pi}, we verify that for every fixed $w$, the policy $\pi_w$ is Lipschitz continuous with respect to $w$ during iteration, and $\pi_w(a|x) > 0$ for all $a \in \mathcal{N}$ and $x \in \mathbb{Z}^N$. iii) Properties of the expectation: We prove that the remaining assumptions are satisfied (Lemma~\ref{lmm: exp bounded}).

\subsection{Stochastic Boundedness and Ergodicity}\label{subsec_MDP_ergodic}
In this section, we prove the boundedness of MGF of the state, then the ergodicity of MDP.

\begin{figure}[htbp]
        \centerline{\includegraphics[width=0.6\linewidth]{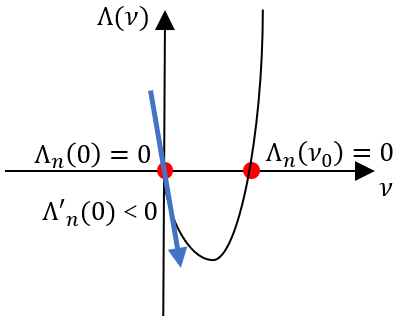}}
    \caption{A parallel service system.}
    \label{fig:lambda_nu_func}
\end{figure}

Before the proof of theorem, we first propose the following lemmas.
It is convenient to define the following auxiliary vector-valued functions $\phi_{(n,j)+},\phi_{(n,j)-}:\mathbb{Z}_{\geq 0}\to\mathbb{R}$ and $ \phi_n,\phi_{n+},\phi_{n-}:\mathbb{Z}_{\geq 0}\to\mathbb{R}^P$ by letting 
\begin{align*}
&\phi_{(n,j)+}(x_n)=\phi_{n,j}(x_n+1)-\phi_{n,j}(x_n),\forall n\in\mathcal{N},\\
&\phi_{(n,j)-}(x_n)=\phi_{n,j}(x_n-1)-\phi_{n,j}(x_n),\forall x_n\geq1,n\in\mathcal{N},\\
&\phi_n(x_n)=\{\phi_{n,j}(x_n)\}_{j=1}^P,\forall n\in\mathcal{N}\\
&\phi_{n+}(x_n)=\{\phi_{(n,j)+}(x_n)\}_{j=1}^P,\forall n\in\mathcal{N}\\
&\phi_{n-}(x_n)=\{\phi_{(n,j)-}(x_n)\}_{j=1}^P,\forall x_n\geq1,n\in\mathcal{N},
\end{align*}
which represent the change of the value of basis function. Define
\begin{align*}
    w_n=[w_{n,1},w_{n,2},\ldots,w_{n,P}]^T,
\end{align*}
and
\begin{align*}
    g(x)=\sum\nolimits_{n=1}^Ne^{\nu w^T_n\phi_{n+}(x_n)},
\end{align*}
where $\nu$ is a positive constant. Thus we have $\mathsf E[g(x)]$ is the MGF of state.
\begin{lem}
    \label{lem: bounded and ergodic} 
    Under the constraints in Theorem~\ref{thm: joint convergence}, let $W_e(x)=\sum_{n=1}^Ne^{\nu w^T_n \phi_n(x_n)},~\nu>0$, then $\forall w\in\mathbb{R}^N$ during iteration, there exists finite non-negative constants $B_w,B_e$ satisfying 
    \begin{align}
    \label{equa: W_e drift condition}
    \mathcal{L}_w W_e(x)\leq -(B_e\cdot g(x)+1)+B_w\mathbb{I}_{\{x=0\}}.
    \end{align} 
    Furthermore, we have
    \begin{align*}
        \lim_{K\rightarrow\infty}\frac{1}{K}\sum\nolimits_{k=0}^{K-1}\mathsf E[g(x)]\leq \frac{B_w}{B_e};
    \end{align*}
    The Markov chain induced by $\pi_w$ is ergodic and has a corresponding invariant state distribution $d_w$.
\end{lem}
\noindent\emph{Proof:}
    Let $m=\arg\min_aQ(x,a;w)$ and define $l_n=\mathbb{I}_{\{x_n\geq 1\}}$. Under the (softmax) weighted shortest queue policy, we have
    \begin{align*}
        &\mathcal{L}_w W_{e}(x) = \sum_{a=1}^N\pi_w(a|x)\Bigg\{\sum\limits_{n=1}^Nl_n\mathsf{p}(x'=x-e_n|x,a)\cdot\\
        &\quad\Big[\exp\Big({\nu w^T_n\phi_n(x_n')}\Big)-\exp\Big({\nu w^T_n\phi_n(x_n)}\Big)\Big]\\
        &\quad+\mathsf{p}(x'=x+e_a|x,a)\cdot\\
        &\quad\Big[\exp\Big({\nu w^T_a\phi_a(x_a')}\Big)-\exp\Big({\nu w^T_a\phi_a(x_a)}\Big)\Big]\Bigg\}.
    \end{align*}
    Suppose that $\iota$ is sufficiently small, we have
    \begin{align*}
        &\mathcal{L}_w W_{e}(x)\nonumber\\
        &\leq\sum_{n=1}^N\mu_ne^{\nu w^T_n\phi_{n+}(x_n)}\Big[\exp\Big(\nu w^T_n(\phi_n(x_n-1)\nonumber\\
        &\quad-\phi_{n+}(x_n))\Big)-\exp\Big(\nu w^T_n(\phi_n(x_n)-\phi_{n+}(x_n))\Big)\nonumber\\
        &+\frac{\lambda}{\sum_n\mu_n} \Big(\exp\Big(\nu w^T_a(\phi_a(x_a+1)-\phi_{a+}(x_a))\Big)\nonumber\\
        &\quad-\exp\Big(\nu w^T_a(\phi_a(x_a)-\phi_{a+}(x_a))\Big)\Big)\Big]+B_w\mathbb{I}_{\{x=0\}}\nonumber\\
        &=\sum_{n=1}^N\mu_ne^{\nu w^T_n\phi_{n+}(x_n)}\cdot\Lambda_n(\nu)+B_w\mathbb{I}_{\{x=0\}},
    \end{align*}
    where $B_w\mathbb{I}_{\{x=0\}}=\sum_{n=1}^N(l_n-1)$. 
    Note that $\Lambda_n(0)=0,~\Lambda_n(\infty)\rightarrow{\infty}$. The derivate of $\Lambda_n(\nu)$ at $\nu=0$ is calculated as
    \begin{align*}
        &\frac{d\Lambda_n}{d\nu}\Bigr|_{\nu=0} =w^T_n\phi_{n-}(x_n)+\frac{\lambda w^T_a}{\sum\mu_n}\phi_{a+}(x_a)
        \\
        &\leq \sum_{j=1}^P w_{n,j} \Big[\Big(\phi_{n,j}(x_n-1)-\phi_{n,j}(x_n)\Big)\nonumber\\
        &\quad\quad+\frac{\lambda}{\sum_n \mu_n}\Big(\phi_{n,j}(x_n+1)-\phi_{n,j}(x_n)\Big)\Big].\nonumber
    \end{align*}
     By Taylor's theorem, we have 
    \begin{align*}
        &\phi_{n,j}(x_n+1)=\phi_{n,j}(x_n)+\phi'_{n,j}(x_n+\epsilon_{n,j}),\quad \epsilon_{n,j}\in(0,1)\\
        &\phi_{n,j}(x_n)=\phi_{n,j}(x_n-1)+\phi'_{n,j}(x_n-1)+R_{n,j},R_{n,j}\geq0,
    \end{align*}
    where $R_{n,j}$ is the remainder of Taylor's formula. Thus we have
    \begin{align*}
        &\frac{d\Lambda_n}{d\nu}\Bigr|_{\nu=0}\\
        &\leq\sum_{j=1}^Pw_{n,j}\Big[-\phi'_{n,j}(x_n-1)+\frac{\lambda}{\sum_n \mu_n}\phi'_{n,j}(x_n+\epsilon_{n,j})\Big]\\
        &\leq\sum_{j=1}^Pw_{n,j}\Big[\Big(\frac{\lambda}{\sum_n \mu_n}-1\Big)\phi'_{n,j}(x_n)\\
        &\quad+\Big(\phi'_{n,j}(x_n)-\phi'_{n,j}(x_n-1)\Big)\\
        &\quad+\frac{\lambda}{\sum_n \mu_n}\Big(\phi'_{n,j}(x_n+\epsilon_{n,j})-\phi'_{n,j}(x_n)\Big)\Big].
    \end{align*}
Again, by Taylor's theorem, we have
\begin{align*}
    &\phi'_{n,j}(x_n)-\phi'_{n,j}(x_n-1)\\
    &=\phi''_{n,j}(x_n)+\frac{1}{2}\phi'''(\xi_{n,j}),\xi_{n,j}\in(x_n-1,x_n),\\
    &\phi'_{n,j}(x_n+\epsilon_{n,j})-\phi'_{n,j}(x_n)\\
    &=\epsilon_{n,j}\phi''_{n,j}(x_n)+\frac{\epsilon_{n,j}^2}{2}\phi'''(x_n+\tilde{\epsilon}_{n,j}),\tilde{\epsilon}_{n,j}\in(0,\epsilon_{n,j}).
\end{align*}
Recall that $\phi_{n,H}(x_n)$ is the function with the highest-degree of $x_n$, thus we have $\lim_{h\rightarrow0}\frac{\phi_{n,i}(x_n)}{\phi_{n,H}(x_n)}=0,i\not=H,i\leq P$.
Then there exists a finite constant $B_{n,H}$ such that for all $w$ 
\begin{align*}
    &\lim_{x_n\rightarrow\infty}\frac{d\Lambda_n(\nu)}{d\nu}\Bigr|_{\nu=0}\\
    &\leq B_{n,H}\Big[\Big(\frac{\lambda}{\sum\mu_n}-1\Big)\phi'_{n,H}(x_n)+2\Big(\frac{\epsilon_{n,H}\lambda}{\sum\mu_n}+1\Big)B_{de}\Big],
\end{align*}
which is negative when the queuing system is stabilizable and Assumption~\ref{assum: basis func} is satisfied.
The negativity of $\frac{d\Lambda_n(\nu)}{d\nu}\Bigr|_{\nu=0}$ implies that there exist $\nu_0>0$ as the second zero of $\Lambda_n(\nu)$ and $\Lambda_n(\nu)<0,~\nu\in(0,\nu_0)$, which is demonstrated in Figure.~\ref{fig:lambda_nu_func}. Now we can conclude that with a proper selection of $\nu$, \eqref{equa: W_e drift condition} is guaranteed. 

    Following the proof from \cite{georgiadis2006resource}, summing the inequality over epochs $k\in\{0,\ldots,K-1\}$ yields a telescoping series on the left hand side of \eqref{equa: W_e drift condition}, result in
    \begin{align*}
        &\mathsf{E}[W_e(x[K])]-\mathsf{E}[W_e(x[0])]\nonumber\\
        &\leq KB_w-\sum_{k=0}^{K-1} \mathsf E\Big[B_e\sum\nolimits_{n=1}^Ne^{\nu w^T_n\phi_{n+}(x_n)}\Big].
     \end{align*}
    Since $\mathsf{E}[W_e(x[K])]\geq0$, we have
    \begin{align*}
        &\lim_{K\rightarrow\infty}\frac{1}{K}\sum_{k=0}^{K-1} \mathsf E\Big[\sum\nolimits_{n=1}^Ne^{\nu w^T_n\phi_{n+}(x_n)}\Big]\\
        &\leq \frac{B_w}{B_e}+\lim_{K\rightarrow\infty}\frac{\mathsf{E}[W_e(x[0])]}{KB_e}= \frac{B_w}{B_e}.
    \end{align*}
    The above inequality implies the boundedness of MGF of states, thus the higher order stability of system states \cite[Proposition 8.2.4]{dai1995stability} and \eqref{equa: stable state}. 
    Then with Foster-Lyapunov criterion \cite[Theorem 4.3.]{meyn_tweedie_1993} and \eqref{equa: W_e drift condition} we further have for all $k,t\geq 0$
    \begin{align}
        \label{equa: exp of W_e}
        \mathsf E\Big[W_e(x[k+t])\Big|x[k]=x,w\Big]\leq W_e(x),x\not=0.
    \end{align}

To argue for the irreducibility of the chain, note that the state $x=0$ can be accessible from any initial condition with positive probability. The policy satisfies $\pi_w(a|x)>0$ for all $x,a$ and $w$. According to \cite[Theorem 14.0.1]{meyn2012markov}, the proof of ergodic and positive Harris recurrent is straightforward with \eqref{equa: W_e drift condition} by letting $V(x)=W_e(x)$ and $f(x)=B_e\cdot g(x)+1$. Thus, there exists an invariant measure $d_w$. 
\hfill$\square$

\subsection{Stochastic Regularity Policy and Basis Function}
Based on the characters of MDP proved in Section~\ref{subsec_MDP_ergodic}, we proposed the following Lemmas to state the qualification of policy and basis function. Specifically, we proved that the expectation of the basis function satisfies specific condition and the weighted shortest queue policy is Lipschistz continuous.
\begin{lem}
\label{lmm: exp bounded}
    Recall that the functions $g_w(x)$ and $r_w(x)$ are defined as \eqref{equa: def TDerror}. There exist constants $B_{qf},q>1$ for all $x\in\mathbb{Z}^n_{\geq0}, w\in\mathbb{R}^{N\cdot P},k\geq 0$ during iteration
    \begin{align*}
        \sum_{t=0}^\infty&\Big\|\sum_{x'}\mathsf p_k^t(x'|x) g_w(x')-\sum_{x'}d_{w}(x')g_w(x')\Big\|_2\\
        &<B_{qf}\big(W^q_e(x)+1\big),\\
        \sum_{t=0}^\infty&\Big\|\sum_{x'}\mathsf p_k^t(x'|x) r_w(x')-\sum_{x'}d_{w}(x')r_w(x')\Big\|_2\\
        &<B_{qf}\big(W^q_e(x)+1\big),\quad x[k]=x,w[k]=w,
    \end{align*}
    where $\mathsf p^t_k(x'|x)$ indicates the transition probability from state $x$ to $x'$ after $t$ steps under fixed policy $\pi_{w}$, and there is $\mathsf p(x'|x)=\sum_{a=1}^N\pi(a|x)\mathsf p(x'|a,x)$. Besides, there exist constants $B_{q}$ such that for all $x,w,k,t$
    \begin{align*}
        \mathsf E[W^q_e(x[k+t])|x[k]=x,w]\leq B_{q}(1+W^q_e(x)).
    \end{align*}
\end{lem}
\noindent\emph{Proof:} We define $g^q(x)$ as
\begin{align*}
    g^q(x) = \Big(\sum\nolimits_{n=1}^Ne^{\nu w^T_n\phi_{n+}(x_n)}\Big)^q.
\end{align*}
Note that by Jensen's inequality, there exist constants $B_{Nq}\leq N^{q-1}$ for all $N,q>1$ such that
\begin{align*}
    g^q(x)&=B_{Nq}\sum\nolimits_{n=1}^N\Big(e^{\nu w^T_n\phi_{n+}(x_n)}\Big)^q\\
    &=B_{Nq}\sum\nolimits_{n=1}^Ne^{q\nu w^T_n\phi_{n+}(x_n)}.
\end{align*}
Then we define $W_e^q(x)$ as
\begin{align*}
    W_e^q(x) = \Big(\sum_{n=1}^Ne^{\nu w^T_n \phi_n(x_n)}\Big)^q,~\nu>0.
\end{align*}
Similarly, there exist constants $B'_{Nq}$ such that $W_e^q(x) = B'_{Nq}\sum_{n=1}^Ne^{q\nu w^T_n \phi_n(x_n)}$. Note that constants $B_{Nq},B'_{Nq},q$ are not related to state, thus \eqref{equa: W_e drift condition} still holds with functions $W_e^q(x)$ and $g^q(x)$. That is, there exist finite positive constants $B_{qe},B_{qw}$ such that for all $q>1$
\begin{align*}
    \mathcal{L}_w W^q_e(x)\leq -(B_{qe}\cdot g^q(x)+1)+B_{qw}\mathbb{I}_{\{x=\mathbf{0}\}}.
\end{align*}
Since we have $\|g_w(x)\|_2=\mathcal{O}\Big(\phi(x,a)\phi(x,a)^T\Big)$, $\|r_w(x)\|_2=\mathcal{O}\Big(\phi(x,a)\cdot\sum_{n=1}C_n(x_n)\Big)$, there exist constants $\tilde{q}>1$ and $B'_{\tilde{q}e}>0$ such that for all $x\in\mathbb{Z}^N_{\geq 0},k\geq 0,w$
\begin{align*}
    \|g_w(x)\|_2, \|r_w(x)\|_2<B'_{\tilde{q}e}g^{\tilde{q}}(x)+1\quad=\quad f_g^{\tilde{q}}(x),
\end{align*}
where $\|\cdot\|_2$ is also used to denote the Euclidean-induced norm on matrix $g_w(x)$.
According to \cite[Theorem 14.0.1]{meyn2012markov}, with the proof of \eqref{equa: W_e drift condition}, we have 
    \begin{align*}
        \sup_{f_p:|f_p|\leq f_g^{\tilde{q}}}&\sum_{t=0}^\infty\Big|\sum_{x'}\mathsf p_k^t(x'|x)f_p(x')-\sum_{x'}d_{w[k]}(x')f_p(x')\Big|\nonumber\\
        &<B'_{qf}(W^{\tilde{q}}_e(x)+1),\quad x[k]=x,
    \end{align*}
    where $B'_{qf}$ is a finite positive constant. Analogous to \eqref{equa: exp of W_e} in Lemma~\ref{lem: bounded and ergodic}, we have that there exist constants $B_{\tilde{q}}>0$ such that for all $k\geq 0$
    \begin{align*}
        \mathsf E\Big[W^{\tilde{q}}_e(x[k])\Big|x[0]=x,w\Big]<B_{\tilde{q}}\Big(W^{\tilde{q}}_e(x)+1\Big).
    \end{align*}
\hfill$\square$

According to \cite[Theorem 2]{Tsitsiklis97}, the proof of Lemma~\ref{lmm: exp bounded} ensures the assumptions of basis function. Besides, we need to analyze the Lipschitz continuity of the policy, which is crucial for the convergence analysis.
\begin{lem}(Lipschitz continuity of $\pi_w(a|x)$) 
\label{lmm:Lip pi} For our policy, there exists $L_\pi>0$ such that $\forall w,w',a,x$,
$$
\sup_{x,a}|\pi_w(a|x)-\pi_{w'}(a|x)|\leq L_{\pi}\|w-w'\|_2.
$$
\end{lem}
\noindent\emph{Proof:}
    Note that the boundedness of derivative towards $w$ implies the Lipschitz continuity. According to \eqref{equa: policy}, we have
    \begin{align*}
        \pi_w(a|x)&=\Big(\sum_{b=1}^N \exp\Big(\frac{1}{\iota}\Big(w^T_a\phi_{a+}(x_a)-w^T_b\phi_{b+}(x_b)\Big)\Big)\Big)^{-1}\\
        &=\frac{1}{\Sigma_a}.
    \end{align*}
    We have the derivation
    \begin{align*}
    &\frac{\partial \pi_w(a|x)}{\partial w_{n,j}}=\\
    &\begin{cases}
        \frac{-\phi_{(a,j)+}(x_a)}{\iota\Sigma_a}&n=a,\\
        \frac{\phi_{(n,j)+}(x_n)\cdot\exp\Big(\frac{1}{\iota}\Big(w^T_a\phi_{a+}(x_a)-w^T_n\phi_{n+}(x_n)\Big)\Big)}{\iota\Sigma_a^2}&n\not=a.
    \end{cases}
\end{align*}
Then we have 
\begin{align*}
    |\frac{\partial \pi_w(a|x)}{\partial w_{n,j}}|\leq\frac{\phi_{(n,j)+}(x_n)}{\iota\Sigma_a}.
\end{align*}
Note that $\Sigma_a\geq1$ and $\lim_{x_n\rightarrow\infty}\frac{\phi_{(n,j)+}(x_n)}{\Sigma_a}=0$. Then we can conclude the boundedness of $\|\frac{d \pi_w(a|x)}{d w}\Big\|_2$, thus the Lipschitz continuity of the policy.
\hfill$\square$

\subsection{Proof of Theorem~\ref{thm: joint convergence}}
With the proof of the above lemmas, we are ready to prove the main result of the paper. We show that the learning policy is Lipschitz continuous w.r.t. weight $w$, $\pi_w(a|x)>0$; The induced chain and basis function satisfy the required assumptions of \cite[Theorem 17]{Benveniste90}, which provides essential assumptions and conclusions of using ODE method for convergence analysis of stochastic approximation algorithms. Therefore, leveraging the results in \cite{melo2008analysis}, we can analyze the convergence of the weight vector $w$ by studying the stability of the associated ordinary differential equation (ODE):
\begin{align}
\label{equa:ODE}
    \dot{w}=\mathsf E_{d_w}\Big[\frac{\alpha_k}{B_\alpha[k]}\delta_{w}(x,w)\Big],
\end{align}
where $\mathsf E_{d_w}[\cdot]$ is the expectation taken with respect to the invariant measure $d_w(\cdot)$ of the chain and initial state $x$, under policy $\pi_w$.
Let's define
\begin{align*}
    \bar \delta_w(x)&=\mathsf E_{d_w}[\delta_w(x,w)]\\
    &=\mathsf E_{d_w}\Big[\phi(x)\Big(\gamma\phi(x')-\phi(x)\Big)^T\Big]\cdot w+\mathsf E_{d_w}\Big[\phi(x)c\Big]\\
    &=\bar g_w(x)\cdot w+\bar r_w(x),
\end{align*}
where $\bar g_w(x)\in\mathbb{R}^{N\times N}$ is a matrix, $\bar r_w(x)\in\mathbb{R}^N$ is a vector. According to \cite{Tsitsiklis97}, with Lemma~\ref{lem: bounded and ergodic}, the basic SARSA(0) has a unique solution $w^*$ satisfies $\bar\delta_{w^*}(x)=0$. 
We show that $w^*$ satisfies \eqref{equa:ODE} and the Bellman equation.
Then we have
\begin{align}
\label{equa: w derivative}
    \frac{d\|w-w^*\|^2_2}{dt}&=2(w-w^*)^T\cdot\dot{w}\nonumber\\
    &=2(w-w^*)^T\frac{\alpha_k\bar\delta_{w}(x)}{B_\alpha[k]}
\end{align}
When the Lipschitz constant $\L_\pi$ is sufficiently small, there exist a $\gamma\in(0,1)$ and a positive constant $B_{d}$ such that $\bar g_w(x)$ is negative definite and 
\begin{align}
\label{equa: w derivative part 1}
    &2(w-w^*)^T\bar\delta_w(x)\nonumber\\
    &=2(w-w^*)^T\Big\{\bar g_w(x)(w-w^*)+\nonumber\\
    &\Big(\bar g_w(x)-\bar g_{w^*}(x)\Big)w^*+(\bar r_w(x)-\bar r_{w^*}(x))\Big\}\nonumber\\
    &\leq 2\Big(-B_A+L_\pi(B_A\|w^*\|_2+B_r)\Big)\|w-w^*\|^2_2\nonumber\\
    &=-2B_{d}\|w-w^*\|^2_2,
\end{align}
where $B_A, B_r$ are finite positive constants that
\begin{align*}
    &-B_A\geq \mathsf E\Big[\Big\|\phi(x)\Big(\gamma\phi(x')-\phi(x)\Big)^T\Big\|_2\Big],\\
    &B_r\geq\mathsf E\Big[\|\phi(x)\cdot c\|_2\Big].
\end{align*}
The above constants are proved to be existed by Lemma~\ref{lem: bounded and ergodic}. 
Thus we have 
\begin{align*}
    \|w[k]-w^*\|^2_2\leq \|w[0]-w^*\|^2_2\cdot e^{-B_d k},
\end{align*}
which ensures the existence of a globally asymptotic equilibrium point of \eqref{equa:ODE}. Recall that $w^*$ is the optimal solution and $\bar\delta_{w^*}(x)=0$, then finally we can derive the almost surely convergence of weight $w$ to the optimal weight $w^*$.
\hfill$\square$

\section{Simulation-based validation}\label{sec_exp}
To show the benefits of the semi-gradient SARSA (SGS) algorithm with weighted shortest queue policy, we consider the congestion control problem in data network. According to \cite{low2002understanding}, the TCP Vegas addresses the problem of maximization of a strictly concave increasing utility function $U(x_s)$ by adjusting the size of transmission window (i.e. the number of packets to be transmitted) $w_s(t)$ thus the source rate $x_s(t)$. Specifically, $U(x_s)$ and $x_s$ are defined as
\begin{align*}
    U(x_s)&=\alpha_sd_s\log(x_s),\\
    x_s(t)&= \frac{w_s(t)}{d_s+q_s(t)},
\end{align*}
where $\alpha_s$ is a artificially designed constant, $d_s$ is the round trip time associated to the basic property of link and $q_s(t)$ is the queuing delay over the link at time $t$. To reformulate the problem, since each job is routed instantly upon arrival, we can set the transmission window of the source node $S$ as $w_s(t)=\lambda$, which is a constant. Thus, we can minimize $q_s(t)$ to maximize the utility function. By Little's Law, we have $q_s(t) = \mathcal{O}(\|x(t)\|_1/\mu)$, where $x(t)$ is the queuing length and $\mu$ is the service rate of the network. Then we have $U(x)=\mathcal{O}(\log\frac{1}{ \|x(t)\|_1})=\mathcal{O}(-\log( \|x(t)\|_1))$, thus the maximization of utility function is equal to the minimization of $\log(x(t))$.
Then we define the cost function following the form of the utility function 
 \begin{align*}
    c[k+1]=\log(\|x[k+1]\|_1)\cdot(t_{k+1}-t_k).
\end{align*}
To approximate the global time-accumulated utility, we consider the basis function as
\begin{align*}
    &\phi_{n,1}(x_n) = 1+x_n^{0.01},\\
    &\phi_{n,2}(x_n) = x_n^{0.2},\\
    &\phi_{n,3}(x_n) = x_n,\\
    &\phi_{n,4}(x_n) = x_n^{1.5},n=1,2,\ldots,N,
\end{align*}
which satisfies Assumption~\ref{assum: basis func}. Note that $\phi_{n,1}(x_n)$ is defined as a monomial of $x_n$ with an extra constant to address the idling condition.

To evaluate the performance of the semi-gradient SARSA (SGS) algorithm with weighted shortest queue policy, we consider the following performance metrics:
\begin{enumerate}
    \item Computational efficiency: training time and number of iterations (if applicable).
    \item Implementation efficiency: the average state value of all states during training (if applicable) and average cost (i.e. the average utility).
\end{enumerate}

For comparison experiments, we consider two benchmarks:
\begin{itemize}
    \item Neural network SARSA (NNS): We constructed a NN for approximation of $Q(x,a)$. The control policy and cost are defined the same as in SGS, except that the weight update is replaced by NN update with adaptive moment estimation (Adam) algorithm \cite{Kingma2014AdamAM}. Specifically, the NN has two fully connected layer with a rectified linear unit (ReLU) activation function. The parameters of the NN are initialized following a normal distribution.
    The loss function is the mean square error between the one-step predicted and calculated state-action-value. Since an exact optimal policy of the original MDP is not readily available, the policy computed by NN is used as an approximate optimal policy.
    \item Join the shortest queue (JSQ) policy: For routing decisions under JSQ policy, we simply select the path with the shortest queue length, that is $a_{JSQ}=\arg \min_{n\leq N} x_n$.
\end{itemize}
To implement our algorithm, consider the congestion control problem over the parallel network in Figure~.\ref{fig:parallelQN}. Suppose that there are $N=3$ parallel servers with the service rate $\mu_1 = 0.5,~\mu_2 = 2.5,~\mu_3=5$ and the arrival rate $\lambda = 2$, all in unit sec$^{-1}$. The policy temperature parameter is set as $\iota=0.01$. The weight is initialized randomly following a uniform distribution. For simulation, a discrete time step of 0.1 sec is used. Since the optimal state value and steady state distribution are unique, we train NNS for $3.0\times10^6$ episodes and then simulate for $10000$ times to obtain a set of 10000 sampled states which follows the steady state distribution. For SGS, we train it for $2.0\times10^6$ episodes. To evaluate the algorithm and the iteration process, we calculate the normalized average state value with the sampled data to demonstrate the performance during training. Besides, we evaluate the three model for $1\times10^6$ episodes to obtain the average utility and use it for performance comparison. All experiments were implemented using 13th Gen Intel(R) Core(TM) i7-13700KF, 3400Mhz with 16.0 GB memory. The results are as follows. 
\begin{figure}[htbp]
        \centerline{\includegraphics[width=1.1\linewidth]{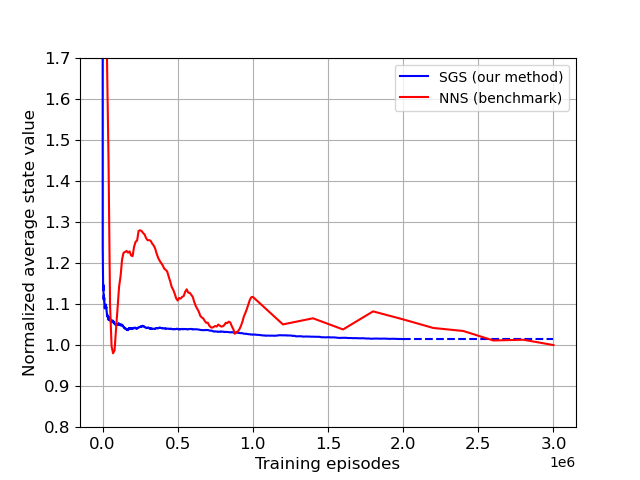}}
    \caption{The convergence performance comparison between SGS and NNS.}
    \label{fig:comp_performance}
\end{figure}

For the training process, the queuing system is first initialized with random queuing length and jobs for both NNS and SGS. We can see from Fig.~\ref{fig:comp_performance} that SGS converges rapidly for several episodes while NNS oscillate for a long time. The average value is calculated with the steady state distribution and normalized by the sub-optimal average state value. 

The final average state value is $0.7213$ and $0.7109$ learned by SGS and NNS, respectively. While the whole training time is $801.48s$ and $4580.69s$ for SGS and NNS, respectively. Besides, our policy is not restricted to non-idling conditions, since $\hat Q(x,n)=w_{n,1},x=0$, the policy can route the new arriving job according to $w_{n,1}$. Thus the server with higher service rate may be selected even the slower server is empty.

\begin{table}[htbp]
    \centering
    \caption{Empirical performance of various schemes.}
    \label{table:NETDC}
    \begin{tabular}{cccc}
        \toprule  
        Algorithm&Normalized Average Cost\\
        \midrule  
        Neural network SARSA (NNS) &1.00\\
        Semi-gradient SARSA (SGS)&1.02\\
        Join the shortest queue (JSQ)&1.74\\
        \bottomrule  
    \end{tabular}
\end{table}
Table \ref{table:NETDC} lists the value of normalized average total discounted cost under various methods. The results are generated from 1000 simulations with initial states following the steady state distribution. Although NN performs better in long terms of learning as expected, SGS performs better with just a few number of iterations as demonstrated in Fig. \ref{fig:comp_performance}.

According to the definition, the utility is related to the logarithm of the reciprocal of the average system time, thus a lower average cost leads to a higher utility. The job will spend more time going through the queuing network under JSQ policy and achieve an average cost of $1.22$. For our method, though the utility of SGS is slightly worse than that of NNS, SGS provides the best trade-off between computational efficiency and implementation efficiency: the average utility of NNS and SGS are $0.70$ and $0.72$. More importantly, SGS algorithm theoretically ensures the convergence of the optimal routing decision, while NNS might be diverge and let alone the existence of the optimal decision.
\section{Conclusions}\label{sec_con}
In this paper, we propose a semi-gradient SARSA(0) (SGS) algorithm with linear generic value function approximation for dynamic routing over parallel servers. The SGS iteration scheme and weighted shortest queue policy are novelly designed and ensure the stability of system during the iteration. We leverage a Lyapunov approach and an ODE-based method to extend the theoretical analysis of SARSA convergence to non-compact state space and unbounded reward. We show that the almost surely convergence of the weight vector to the optimal in SGS is coupled with the convergence of the traffic state, and the joint convergence is guaranteed if and only if the parallel service system is stabilizable. Specifically, the approximator is used as a Lyapunov function for traffic state stability analysis; and the constraint and convergence analysis of weight vector is based on stochastic approximation theory. We prove the long average boundedness of moment generating function, thus the higher order asymptotic stability of the system and the routing policy with generic approximator.
Additionally, we implement the algorithm to the classic congestion control problem in a data network. We compare the proposed SGS algorithm with a neural network-based algorithm and show that our algorithm converges faster with a higher computationally efficiency and an insignificant optimality gap. Possible future work includes (i) extension of the joint convergence result as well as SGS algorithm to a general service network and (ii) the analysis of approximation error and convergence rate.

\section*{Acknowledgment}

This work was supported by NSFC Project 62473250, SJTU-UM Joint Institute, and J. Wu \& J. Sun Endowment Fund. We appreciate discussions with Yuzhen Zhan, Yu Tang and other members of our Smart \& Connected Systems (SCS) Lab.

% \section*{References}
\bibliographystyle{IEEEtran} 
\bibliography{references24}

\begin{thebibliography}{10}
\providecommand{\url}[1]{#1}
\csname url@samestyle\endcsname
\providecommand{\newblock}{\relax}
\providecommand{\bibinfo}[2]{#2}
\providecommand{\BIBentrySTDinterwordspacing}{\spaceskip=0pt\relax}
\providecommand{\BIBentryALTinterwordstretchfactor}{4}
\providecommand{\BIBentryALTinterwordspacing}{\spaceskip=\fontdimen2\font plus
\BIBentryALTinterwordstretchfactor\fontdimen3\font minus \fontdimen4\font\relax}
\providecommand{\BIBforeignlanguage}[2]{{%
\expandafter\ifx\csname l@#1\endcsname\relax
\typeout{** WARNING: IEEEtran.bst: No hyphenation pattern has been}%
\typeout{** loaded for the language `#1'. Using the pattern for}%
\typeout{** the default language instead.}%
\else
\language=\csname l@#1\endcsname
\fi
#2}}
\providecommand{\BIBdecl}{\relax}
\BIBdecl

\bibitem{mitra1991comparative}
D.~Mitra and J.~B. Seery, ``Comparative evaluations of randomized and dynamic routing strategies for circuit-switched networks,'' \emph{IEEE Transactions on Communications}, vol.~39, no.~1, pp. 102--116, 1991.

\bibitem{down1997piecewise}
D.~Down and S.~P. Meyn, ``Piecewise linear test functions for stability and instability of queueing networks,'' \emph{Queueing Systems}, vol.~27, no. 3-4, pp. 205--226, 1997.

\bibitem{alanyali1997analysis}
M.~Alanyali and B.~Hajek, ``Analysis of simple algorithms for dynamic load balancing,'' \emph{Mathematics of Operations Research}, vol.~22, no.~4, pp. 840--871, 1997.

\bibitem{kumar1995stability}
P.~Kumar and S.~P. Meyn, ``Stability of queueing networks and scheduling policies,'' \emph{IEEE Transactions on Automatic Control}, vol.~40, no.~2, pp. 251--260, 1995.

\bibitem{dai1995stability}
J.~G. Dai and S.~P. Meyn, ``Stability and convergence of moments for multiclass queueing networks via fluid limit models,'' \emph{IEEE Transactions on Automatic Control}, vol.~40, no.~11, pp. 1889--1904, 1995.

\bibitem{Jin2018Stability}
L.~Jin and S.~Amin, ``Stability of fluid queueing systems with parallel servers and stochastic capacities,'' \emph{IEEE Transactions on Automatic Control}, vol.~63, no.~11, pp. 3948--3955, 2018.

\bibitem{dai2022queueing}
J.~G. Dai and M.~Gluzman, ``Queueing network controls via deep reinforcement learning,'' \emph{Stochastic Systems}, vol.~12, no.~1, pp. 30--67, 2022.

\bibitem{xie2022stabilizing}
Q.~Xie and L.~Jin, ``Stabilizing queuing networks with model data-independent control,'' \emph{IEEE Transactions on Control of Network Systems}, vol.~9, no.~3, pp. 1317--1326, 2022.

\bibitem{sutton2018reinforcement}
R.~S. Sutton and A.~G. Barto, \emph{Reinforcement learning: An introduction}.\hskip 1em plus 0.5em minus 0.4em\relax MIT press, 2018.

\bibitem{HanZhang20}
M.~Han, L.~Zhang, J.~Wang, and W.~Pan, ``Actor-critic reinforcement learning for control with stability guarantee,'' \emph{IEEE Robotics and Automation Letters}, vol.~5, no.~4, pp. 6217--6224, 2020.

\bibitem{liu2022rl}
B.~Liu, Q.~Xie, and E.~Modiano, ``Rl-qn: A reinforcement learning framework for optimal control of queueing systems,'' \emph{ACM Transactions on Modeling and Performance Evaluation of Computing Systems}, vol.~7, no.~1, pp. 1--35, 2022.

\bibitem{Camelo20}
M.~Camelo, M.~Claeys, and S.~Latré, ``Parallel reinforcement learning with minimal communication overhead for iot environments,'' \emph{IEEE Internet of Things Journal}, vol.~7, no.~2, pp. 1387--1400, 2020.

\bibitem{hoang2023deep}
L.~T. Hoang, C.~T. Nguyen, and A.~T. Pham, ``Deep reinforcement learning-based online resource management for uav-assisted edge computing with dual connectivity,'' \emph{IEEE/ACM Transactions on Networking}, vol.~31, no.~6, pp. 2761--2776, 2023.

\bibitem{GuptaDRL}
J.~K. Gupta, M.~Egorov, and M.~Kochenderfer, ``Cooperative multi-agent control using deep reinforcement learning,'' in \emph{Autonomous Agents and Multiagent Systems}, G.~Sukthankar and J.~A. Rodriguez-Aguilar, Eds.\hskip 1em plus 0.5em minus 0.4em\relax Cham: Springer International Publishing, 2017, pp. 66--83.

\bibitem{schulman2017proximal}
J.~Schulman, F.~Wolski, P.~Dhariwal, A.~Radford, and O.~Klimov, ``Proximal policy optimization algorithms,'' \emph{arXiv preprint arXiv:1707.06347}, 2017.

\bibitem{ChenMAPPO}
L.~Chen, B.~Hu, Z.-H. Guan, L.~Zhao, and X.~Shen, ``Multiagent meta-reinforcement learning for adaptive multipath routing optimization,'' \emph{IEEE Transactions on Neural Networks and Learning Systems}, vol.~33, no.~10, pp. 5374--5386, 2022.

\bibitem{Quentin24NUDGE}
Q.~Delfosse, H.~Shindo, D.~S. Dhami, and K.~Kersting, ``Interpretable and explainable logical policies via neurally guided symbolic abstraction,'' in \emph{Proceedings of the 37th International Conference on Neural Information Processing Systems}, ser. NIPS '23.\hskip 1em plus 0.5em minus 0.4em\relax Red Hook, NY, USA: Curran Associates Inc., 2024.

\bibitem{Huan22}
H.~Yu, S.~Park, A.~Bayen, S.~Moura, and M.~Krstic, ``Reinforcement learning versus pde backstepping and pi control for congested freeway traffic,'' \emph{IEEE Transactions on Control Systems Technology}, vol.~30, no.~4, pp. 1595--1611, 2022.

\bibitem{Heeswijk2017TheDD}
\BIBentryALTinterwordspacing
W.~van Heeswijk, M.~R.~K. Mes, and J.~M.~J. Schutten, ``The delivery dispatching problem with time windows for urban consolidation centers,'' \emph{Transp. Sci.}, vol.~53, pp. 203--221, 2017. [Online]. Available: \url{https://api.semanticscholar.org/CorpusID:125346370}
\BIBentrySTDinterwordspacing

\bibitem{ULMER2020183}
\BIBentryALTinterwordspacing
M.~W. Ulmer and B.~W. Thomas, ``Meso-parametric value function approximation for dynamic customer acceptances in delivery routing,'' \emph{European Journal of Operational Research}, vol. 285, no.~1, pp. 183--195, 2020. [Online]. Available: \url{https://www.sciencedirect.com/science/article/pii/S0377221719303637}
\BIBentrySTDinterwordspacing

\bibitem{de2000existence}
D.~P. De~Farias and B.~Van~Roy, ``On the existence of fixed points for approximate value iteration and temporal-difference learning,'' \emph{Journal of Optimization Theory and Applications}, vol. 105, pp. 589--608, 2000.

\bibitem{Karmakar21}
P.~Karmakar and S.~Bhatnagar, ``Stochastic approximation with iterate-dependent markov noise under verifiable conditions in compact state space with the stability of iterates not ensured,'' \emph{IEEE Transactions on Automatic Control}, vol.~66, no.~12, pp. 5941--5954, 2021.

\bibitem{yaji2019analysis}
V.~G. Yaji and S.~Bhatnagar, ``Analysis of stochastic approximation schemes with set-valued maps in the absence of a stability guarantee and their stabilization,'' \emph{IEEE Transactions on Automatic Control}, vol.~65, no.~3, pp. 1100--1115, 2019.

\bibitem{melo2008analysis}
F.~S. Melo, S.~P. Meyn, and M.~I. Ribeiro, ``An analysis of reinforcement learning with function approximation,'' in \emph{Proceedings of the 25th International Conference on Machine Learning}, 2008, pp. 664--671.

\bibitem{robbins1951stochastic}
H.~Robbins and S.~Monro, ``A stochastic approximation method,'' \emph{The annals of mathematical statistics}, pp. 400--407, 1951.

\bibitem{Qu20}
\BIBentryALTinterwordspacing
G.~Qu and A.~Wierman, ``Finite-time analysis of asynchronous stochastic approximation and $q$-learning,'' in \emph{Proceedings of Thirty Third Conference on Learning Theory}, ser. Proceedings of Machine Learning Research, J.~Abernethy and S.~Agarwal, Eds., vol. 125.\hskip 1em plus 0.5em minus 0.4em\relax PMLR, 09--12 Jul 2020, pp. 3185--3205. [Online]. Available: \url{https://proceedings.mlr.press/v125/qu20a.html}
\BIBentrySTDinterwordspacing

\bibitem{CHEN2022110623}
\BIBentryALTinterwordspacing
Z.~Chen, S.~Zhang, T.~T. Doan, J.-P. Clarke, and S.~T. Maguluri, ``Finite-sample analysis of nonlinear stochastic approximation with applications in reinforcement learning,'' \emph{Automatica}, vol. 146, p. 110623, 2022. [Online]. Available: \url{https://www.sciencedirect.com/science/article/pii/S0005109822004873}
\BIBentrySTDinterwordspacing

\bibitem{chen2024lyapunov}
Z.~Chen, S.~T. Maguluri, S.~Shakkottai, and K.~Shanmugam, ``A lyapunov theory for finite-sample guarantees of markovian stochastic approximation,'' \emph{Operations Research}, vol.~72, no.~4, pp. 1352--1367, 2024.

\bibitem{Tsitsiklis97}
J.~Tsitsiklis and B.~Van~Roy, ``An analysis of temporal-difference learning with function approximation,'' \emph{IEEE Transactions on Automatic Control}, vol.~42, no.~5, pp. 674--690, 1997.

\bibitem{shah2020stable}
D.~Shah, Q.~Xie, and Z.~Xu, ``Stable reinforcement learning with unbounded state space,'' \emph{arXiv preprint arXiv:2006.04353}, 2020.

\bibitem{grosof2024convergence}
I.~Grosof, S.~T. Maguluri, and R.~Srikant, ``Convergence for natural policy gradient on infinite-state average-reward markov decision processes,'' \emph{arXiv preprint arXiv:2402.05274}, 2024.

\bibitem{cdc24}
Y.~Wu, J.~Zhang, and L.~Jin, ``On joint convergence of traffic state and weight vector in learning-based dynamic routing with value function approximation,'' in \emph{2024 63rd IEEE Conference on Decision and Control (CDC)}, 2024.

\bibitem{meyn_tweedie_1993}
S.~P. Meyn and R.~L. Tweedie, ``Stability of markovian processes iii: Foster–lyapunov criteria for continuous-time processes,'' \emph{Advances in Applied Probability}, vol.~25, no.~3, p. 518–548, 1993.

\bibitem{Benveniste90}
A.~Benveniste, M.~Metivier, and P.~Priouret, \emph{Adaptive Algorithms and Stochastic Approximations}, 1st~ed.\hskip 1em plus 0.5em minus 0.4em\relax Springer Publishing Company, Incorporated, 2012.

\bibitem{gallager2013stochastic}
R.~G. Gallager, \emph{Stochastic Processes: Theory for Applications}.\hskip 1em plus 0.5em minus 0.4em\relax Cambridge University Press, 2013.

\bibitem{meyn2012markov}
S.~P. Meyn and R.~L. Tweedie, \emph{Markov chains and stochastic stability}.\hskip 1em plus 0.5em minus 0.4em\relax Springer Science \& Business Media, 2012.

\bibitem{gurvich2009scheduling}
I.~Gurvich and W.~Whitt, ``Scheduling flexible servers with convex delay costs in many-server service systems,'' \emph{Manufacturing \& Service Operations Management}, vol.~11, no.~2, pp. 237--253, 2009.

\bibitem{mandelbaum2004scheduling}
A.~Mandelbaum and A.~L. Stolyar, ``Scheduling flexible servers with convex delay costs: Heavy-traffic optimality of the generalized c$\mu$-rule,'' \emph{Operations Research}, vol.~52, no.~6, pp. 836--855, 2004.

\bibitem{dai2008optimal}
J.~G. Dai and T.~Tezcan, ``Optimal control of parallel server systems with many servers in heavy traffic,'' \emph{Queueing Systems}, vol.~59, no.~2, pp. 95--134, 2008.

\bibitem{singh2000convergence}
S.~Singh, T.~Jaakkola, M.~L. Littman, and C.~Szepesv{\'a}ri, ``Convergence results for single-step on-policy reinforcement-learning algorithms,'' \emph{Machine learning}, vol.~38, pp. 287--308, 2000.

\bibitem{georgiadis2006resource}
L.~Georgiadis, M.~J. Neely, L.~Tassiulas \emph{et~al.}, ``Resource allocation and cross-layer control in wireless networks,'' \emph{Foundations and Trends{\textregistered} in Networking}, vol.~1, no.~1, pp. 1--144, 2006.

\bibitem{low2002understanding}
S.~H. Low, L.~L. Peterson, and L.~Wang, ``Understanding tcp vegas: a duality model,'' \emph{Journal of the ACM (JACM)}, vol.~49, no.~2, pp. 207--235, 2002.

\bibitem{Kingma2014AdamAM}
D.~P. Kingma and J.~Ba, ``Adam: A method for stochastic optimization,'' \emph{arXiv preprint arXiv:1412.6980}, 2014.

\end{thebibliography}

\end{document}